\documentclass[runningheads]{llncs}

\usepackage{eccv}

\usepackage{array}    
\usepackage{graphicx} 
\usepackage{subcaption}

\usepackage{adjustbox} 
\usepackage{xcolor}
\definecolor{eccvblue}{RGB}{0, 0, 255} 

\usepackage{eccvabbrv}

\usepackage{graphicx}
\usepackage{booktabs}
\usepackage{multirow}

\usepackage[accsupp]{axessibility}  

\usepackage[pagebackref,breaklinks,colorlinks,citecolor=eccvblue]{hyperref}

\usepackage{orcidlink}

\begin{document}

\title{
High-Order Evolving Graphs for Enhanced Representation of Traffic Dynamics
}

\titlerunning{HEG for Traffic Dynamics}

 \author{Aditya Humnabadkar\inst{1}\orcidlink{0000-0002-9301-393X} \and Arindam Sikdar\inst{2}\orcidlink{0000-0002-5697-0060} \and
Benjamin Cave\inst{3}\orcidlink{0009-0007-7999-4595} \and Huaizhong Zhang\orcidlink{0000-0001-7867-9453} \and Paul Bakaki\orcidlink{0000-0001-8277-2554} \and Ardhendu Behera\inst{1}\orcidlink{0000-0003-0276-9000}
}

\authorrunning{A. Humnabadkar, A. Sikdar, \etal}

\institute{Department of Computer Sciences, Edge Hill University, Ormskirk, United Kingdom }

\maketitle

\begin{abstract}
We present an innovative framework for traffic dynamics analysis using High-Order Evolving Graphs, designed to improve spatio-temporal representations in autonomous driving contexts. Our approach constructs temporal bidirectional bipartite graphs that effectively model the complex interactions within traffic scenes in real-time. By integrating Graph Neural Networks (GNNs) with high-order multi-aggregation strategies, we significantly enhance the modeling of traffic scene dynamics, providing a more accurate and detailed analysis of these interactions. Additionally, we incorporate inductive learning techniques inspired by the GraphSAGE framework, enabling our model to adapt to new and unseen traffic scenarios without the need for retraining, thus ensuring robust generalization. Through extensive experiments on the ROAD and ROAD Waymo datasets, we establish a comprehensive baseline for further developments, demonstrating the potential of our method in accurately capturing traffic behavior. Our results emphasize the value of high-order statistical moments and feature-gated attention mechanisms in improving traffic behavior analysis, laying the groundwork for advancing autonomous driving technologies. Our source code is available at: \url{https://github.com/Addy-1998/High\_Order\_Graphs}

\end{abstract}

\section{Introduction}
\vspace{-0.1cm}
\label{sec:intro}
Recognizing driving activities and actions from video and image data is a crucial task in computer vision and pattern recognition, particularly for enhancing the safety of autonomous vehicles and preventing collisions. This technology is integral to Advanced Driver Assistance Systems (ADAS), contributing to the development of features such as adaptive cruise control and lane-keeping assistance, which are essential for maintaining safe vehicle distances and providing timely alerts for lane changes~\cite{yurtsever2020survey}. The complexity of driving scene-based activity detection arises from the highly dynamic and unpredictable nature of traffic environments, especially in urban areas where the overall scene dynamics, including the movement and interaction of various road users such as vehicles, pedestrians, and cyclists, play a critical role~\cite{chougule2023comprehensive,chen2024end}. These scene dynamics are influenced by factors such as traffic signals, road conditions, and the collective behaviors of road users, making it challenging to accurately predict and recognize activities~\cite{mishra2023sensing}. The diversity in driving behaviors and techniques across different regions and countries further complicates this task, highlighting the need for efficient algorithms capable of accurately interpreting a wide range of driving actions~\cite{thomas2020perception}. Addressing these challenges is vital for improving traffic safety and enhancing the effectiveness of autonomous driving technologies. \\
\begin{figure}[t!]
    \vspace{-0.1cm}
    \centering
    \begin{subfigure}[b]{0.46\linewidth}
        \centering
        \includegraphics[width=\linewidth]{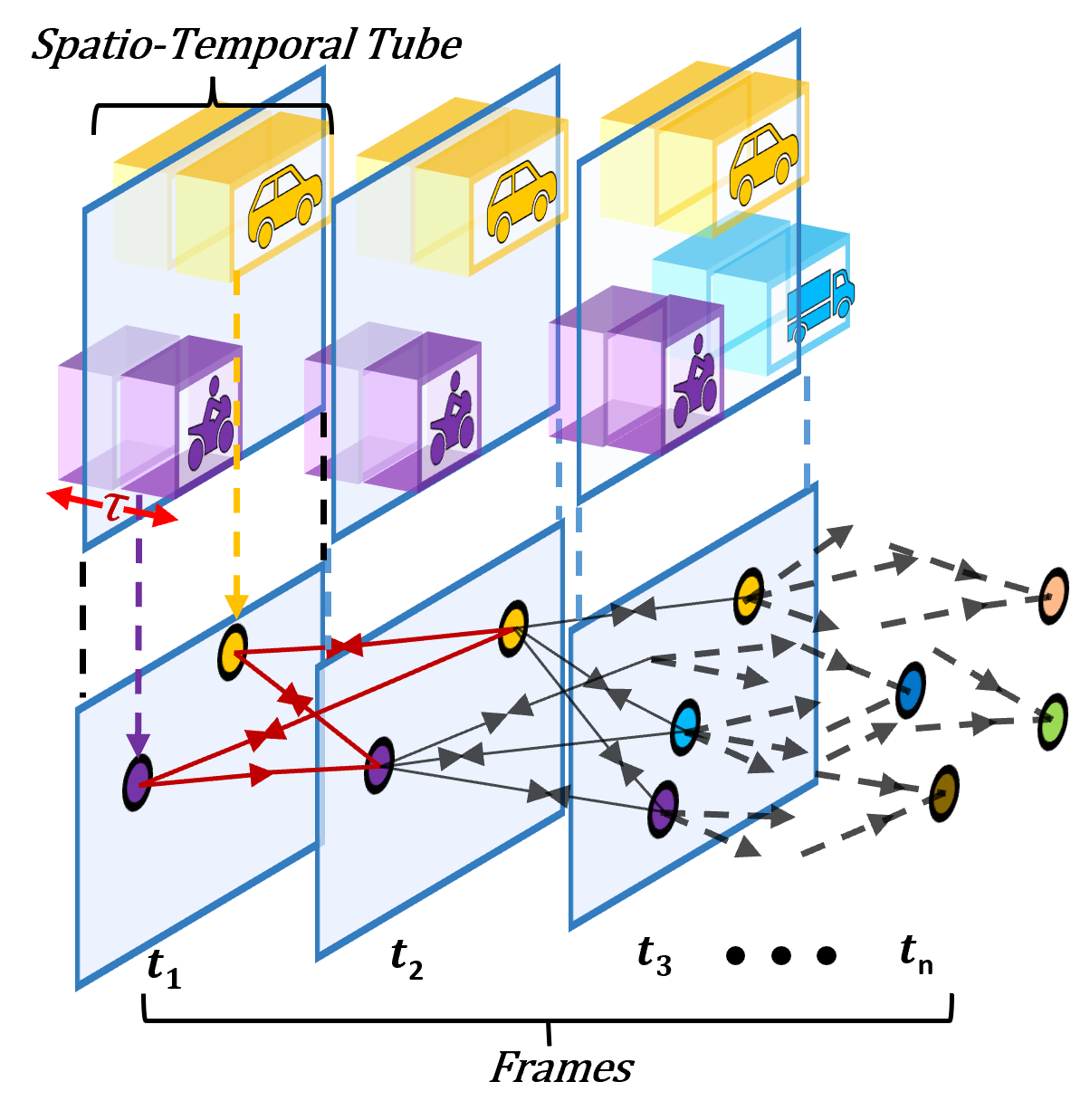}
        \caption{Graph Construction.}
        \label{fig:eccv}
    \end{subfigure}
    \hfill
    \begin{subfigure}[b]{0.45\linewidth}
        \centering
        \includegraphics[width=\linewidth]{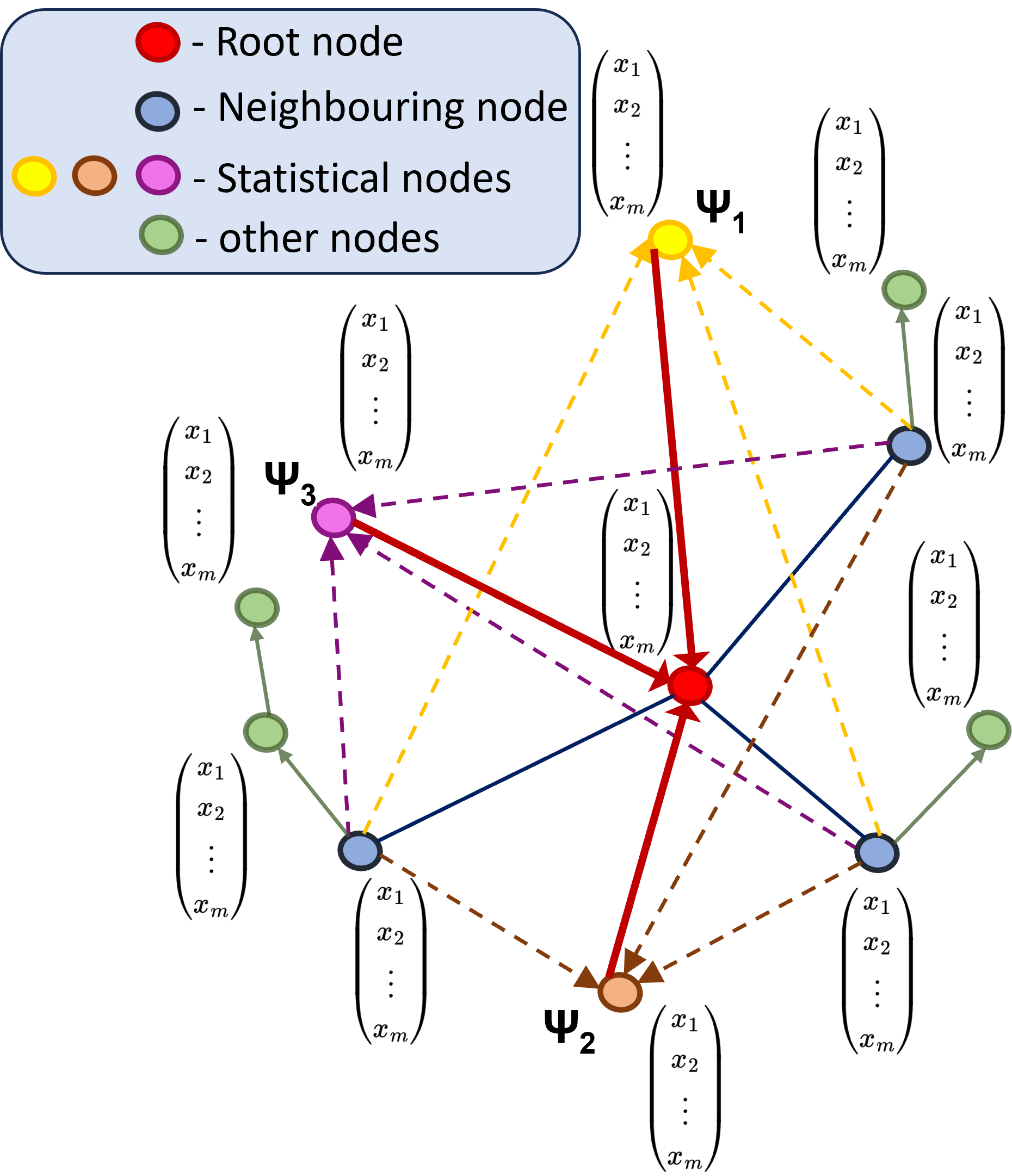}
        \caption{Inductive learning approach}
        \label{fig:graph_sage}
    \end{subfigure}
    \caption{(a) Temporal bidirectional bipartite graph construction from spatio-temporal tube features. (b) Inductive learning framework updating root node features via generated nodes through statistical aggregation from neighbors.}
    \label{fig:side_by_side_graphs}
    \vspace{-0.3cm}
\end{figure}
\indent Traditional video activity recognition methods primarily focus on capturing the spatio-temporal dynamics of objects by tracking features like bounding boxes, 3D coordinates, and key points across video frames to understand how an object's spatial and temporal attributes evolve over time~\cite{behera2014egocentric,jiang2019stm}. While these methods effectively capture broader spatio-temporal aspects, they often struggle with finer details, especially in cluttered environments where subtle movements are crucial~\cite{chappa2023spartan}. To overcome these limitations, deep learning techniques such as Convolutional Neural Networks (CNNs)~\cite{abbas2018video} and Long Short-Term Memory Networks (LSTMs)~\cite{narayanan2019dynamic}, along with more advanced approaches like 3D Convolutional Neural Networks (3D CNNs)~\cite{tran2015learning}, have been adopted. Unlike 2D CNNs that capture only spatial features from individual frames, 3D CNNs process both spatial and temporal data simultaneously, making them more effective at recognizing complex actions in dynamic scenes~\cite{tran2015learning}. However, these models still face challenges, particularly with videos that require an understanding of extended temporal contexts spanning large video durations~\cite{ettinger2021large}. These methods typically process segmented video clips rather than entire sequences, limiting their ability to capture long-term dependencies. They also require substantial computational resources and may struggle to fully capture intricate relationships in dynamic environments, crucial for understanding traffic behavior~\cite{schiappa2023large}. Our approach uses 3D CNNs to extract detailed spatio-temporal features from short object movements within continuous video streams. These features are connected through a our proposed graph structure, enabling efficient long-term video representation while managing computational demands, improving the model's ability to represent complex traffic dynamics. \\
\indent Graph-based approaches~\cite{monninger2023scene,zhou2022toward} have become increasingly popular for tackling the complexities of traffic scene analysis by effectively modeling spatial and temporal relationships between objects. Spatio-temporal traffic graphs, for instance, capture the evolution of traffic states, enabling the identification of congestion patterns and flow dynamics in mixed traffic conditions~\cite{roy2020defining}. Inductive learning techniques in graph-based models have further enhanced these approaches by enabling generalization to unseen data through feature-based embeddings, rather than relying solely on fixed graph structures~\cite{hamilton2017inductive,yang2016revisiting}. This is particularly crucial for real-time traffic analysis, where adaptability to dynamic environments is essential. Additionally, advanced aggregation techniques in Graph Neural Networks (GNNs) boost the representational power of models by effectively integrating information from neighboring nodes while maintaining computational efficiency~\cite{ding2022af2gnn}. By leveraging these advancements, our approach offers a scalable and robust solution for accurate activity recognition in complex driving scenarios. \\
\indent In this work, we present a novel framework that utilizes inductive learning into graph-based approaches for driving scene analysis, inspired by the GraphSAGE~\cite{hamilton2017inductive} framework. We enhance the capabilities of GNNs by introducing a more effective node generation mechanism as part of the inductive learning process, incorporating a multi-aggregation approach~\cite{ding2022af2gnn}. This inductive approach allows our model to adapt to new scenarios without the need for retraining on a fixed graph structure, making it particularly suitable for real-time applications in dynamic and evolving traffic environments. We employ state-of-the-art (SotA) feature extraction techniques to capture robust spatio-temporal representations of objects within a very short time span. By incorporating high-order statistics into the aggregation process, we construct a high-order evolving graph that efficiently captures the intricate details of scene dynamics in video data. This ensures that our model remains accurate and robust, effectively handling the complexity of driving scenarios.

\subsection{Related Work}
\textbf{Spatio-Temporal Approaches.}  
Spatio-temporal methods in video analysis have been instrumental in advancing the understanding of object movements and interactions over time, which is crucial for tackling complex video tasks. Foundational models like Separable 3D CNNs (S3D)~\cite{xie2018rethinking}, R(2+1)D~\cite{tran2018closer}, SlowFast networks~\cite{feichtenhofer2019slowfast}, and 3D ResNet~\cite{hara2018towards} have been key in extracting rich spatio-temporal features. These architectures have improved motion representation and classification accuracy, setting benchmarks in processing efficiency and depth of analysis across various video applications, including action recognition and scene understanding in dynamic environments. Building on these foundations, recent approaches have introduced more sophisticated mechanisms for refining video analysis. For instance, Diba~\etal~\cite{diba2023spatio} proposed a convolution-attention network that balances short and long-range temporal cues using a hierarchical structure, providing nuanced temporal understanding necessary for complex video reasoning. Similarly, Chen~\etal~\cite{chen2023neural} utilized Spatio-Temporal Cross-Covariance Transformers in neural video compression, effectively merging spatial and temporal data with advanced 3D convolutional strategies and attention mechanisms, which are crucial for managing large-scale video data in autonomous driving applications. Li~\etal~\cite{li2023discovering} introduced the Spatio-Temporal Rationalizer (STR), a technique that dynamically identifies crucial video moments and objects, enhancing model performance on multifaceted datasets—an essential capability for accurate, context-aware traffic analysis in autonomous driving. Our work builds on these models as backbones, seamlessly integrating them into our GNN framework to significantly enhance scene classification, action recognition, and scene understanding in autonomous driving. \\
\begin{figure}[!b]
    \vspace{-0.1cm}
    \centering
    \includegraphics[width=0.98\linewidth]{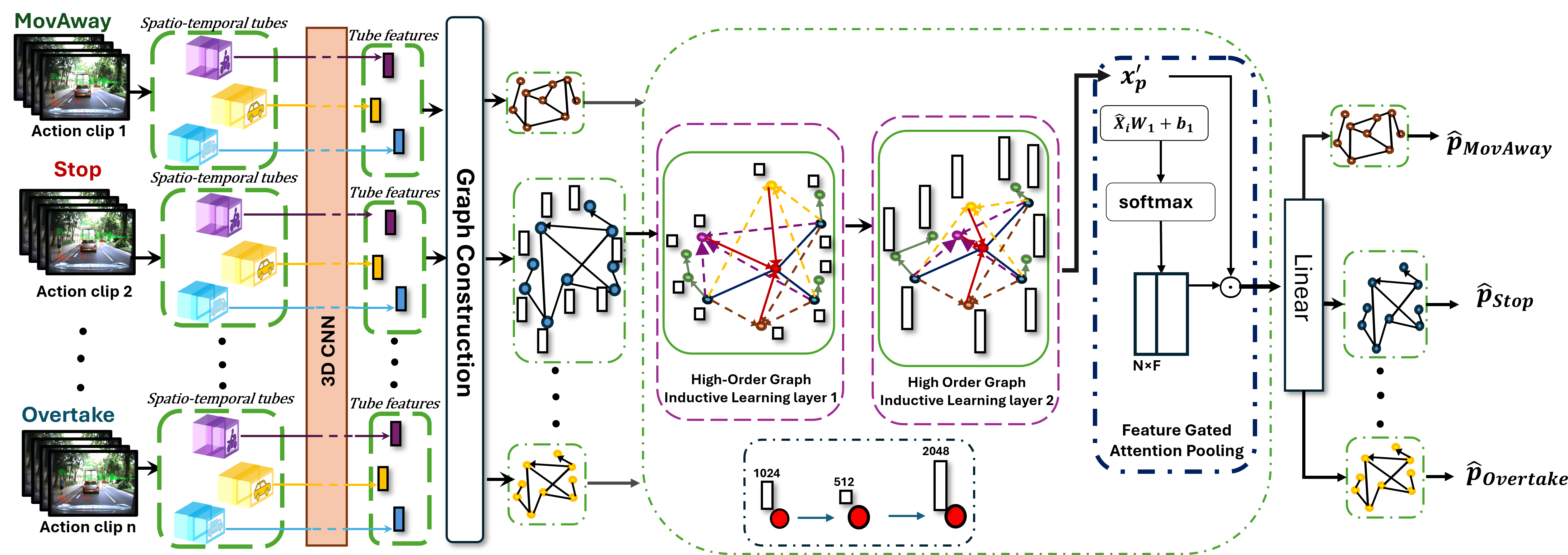}
    \caption{
    The proposed pipeline begins with (a) graph construction, where nodes receive features extracted from spatio-temporal tubes via a 3D CNN. The graph is then processed through (b) high-order multi-aggregation based inductive learning layers, followed by (c) feature-level gated attention aggregation and a fully connected linear layer for classification.}  
    \label{fig:pipeline}
    \vspace{-0.3cm}
\end{figure}
\noindent \textbf{Graph-Based Approaches.}  
Recent advancements in GNNs~\cite{mlodzian2023nuscenes,thakur2024graph,liu2023traffic} for traffic scene analysis have shown substantial progress in traffic management and vehicle behavior prediction. Foundational work includes Visual Traffic Knowledge Graphs by Guo~\etal~\cite{guo2023visual}, which parse complex interactions within traffic scenes to improve semantic understanding. Mlodzian~\etal~\cite{mlodzian2023nuscenes} contributed the nuScenes Knowledge Graph, which provides a comprehensive semantic representation that enhances trajectory prediction by incorporating detailed contextual information like road topology and traffic regulations. Further advancements include Thakur~\etal~\cite{thakur2024graph}, who developed a nested graph-based framework for early accident anticipation, contributing to proactive traffic safety. Li~\etal~\cite{li2020evolvegraph} introduced EvolveGraph, a framework for multi-agent trajectory prediction using dynamic relational reasoning, which significantly improves the understanding and forecasting of agent movements in complex environments. Kumar~\etal~\cite{kumar2021interaction} refined trajectory prediction with an interaction-based model that incorporates agent interactions over a hybrid traffic graph, enhancing accuracy in mixed traffic scenarios. Jin~\etal~\cite{jin2023fast} focused on fast contextual scene graph generation with unbiased context augmentation for dynamic scene understanding, while Hugle~\etal~\cite{hugle2020dynamic} explored dynamic interaction-aware scene understanding crucial for reinforcement learning in autonomous driving. Additionally, Tian~\etal~\cite{tian2024rsg} developed RSG-Search Plus, a traffic scene retrieval method based on road scene graphs, facilitating efficient scene indexing and retrieval. Liu~\etal~\cite{liu2023traffic} combined traffic scenario understanding with video captioning, using a guidance attention captioning network to generate informative video summaries. Wang~\etal~\cite{wang2023gsc} proposed a graph and spatio-temporal continuity framework for accident anticipation, highlighting GNNs' utility in predicting traffic incidents. Sadid~\etal~\cite{sadid2024dynamic} introduced a dynamic spatio-temporal GNN for surrounding-aware trajectory prediction in autonomous vehicles, and Zhou~\etal~\cite{zhou2024hktsg} presented a hierarchical knowledge-guided traffic scene graph representation framework that supports intelligent vehicle applications by learning complex traffic scene dynamics. Büchner~\etal~\cite{buchner2023learning} investigated the aggregation of lane graphs for urban automated driving, enhancing navigational strategies. Unlike these approaches, our method employs high-order evolving graphs based on inductive learning to better capture long-term dependencies and complex interactions, significantly boosting accuracy and robustness in scene understanding and action recognition for autonomous driving.

\vspace{-0.3cm}
\section{Proposed Approach}
\label{sec:proposed_approach}
\vspace{-0.1cm}
Our model pipeline, depicted in Fig.~\ref{fig:pipeline}, classifies events in traffic scenes using a dataset of $N$ videos, denoted as $V_n$ where $n = \{1, 2, \dots, N\}$, each with a ground truth label $p_n$. The objective is to predict class probabilities $\hat{p}_n$ that closely match the true labels $p_n$. The approach comprises three main components: 1) Spatio-temporal tube feature extraction using a 3D CNN backbone, 2) Construction of a temporal bidirectional bipartite graph linking objects across consecutive frames based on extracted features as node features, and 3) A multi-aggregation mechanism incorporates high-order statistics from neighboring nodes. This mechanism utilizes an inductive representation-based graph learning framework~\cite{hamilton2017inductive}. The process is finalized with feature-level gated pooling and a classification layer, producing the class probabilities $\hat{p}_n$ to match the true labels $p_n$ accurately.
\vspace{-0.2cm}
\subsection{Spatio-Temporal Tube Features}
\vspace{-0.1cm}
\label{sub:spatio-temporal-tubes}
We start by generating spatio-temporal tubes for each object in every video frame. For each object a spatio-temporal tube is created based on its bounding box, capturing the object's appearance across a sequence of frames. This tube spans $\tau$ consecutive frames, with $\tau/2$ frames preceding and $\tau/2$ frames following the central frame. The resulting spatio-temporal tube forms a 4D tensor with dimensions of $\tau$ (time) $\times$ $w$ (width) $\times$ $h$ (height) $\times$ $C$ (channels), where $w \times h$ represents the bounding box size of the object in the central frame. We then adjust the spatial resolution of these tensors using bilinear interpolation to match the input dimensions required by the pre-trained 3D CNN model. The S3D model~\cite{xie2018rethinking}, a SotA 3D CNN designed for video classification tasks, is employed as our feature extractor to process these spatio-temporal tubes. It is important to note that this feature extraction serves as a pre-processing step and does not involve any training. The extracted features are used as node features in our bipartite graph, modeling dynamic interactions between consecutive frames.
\vspace{-0.2cm}
\subsection{Temporal Bidirectional Bipartite Graph}
\textbf{Graph Construction.} Given a video $V_n$ consisting of $T$ frames, we construct a multi-frame bidirectional bipartite graph $\mathcal{G}_n(\mathcal{V}_n, \mathcal{E}_n)$ to represent the interactions among objects between consecutive frames. The node set $\mathcal{V}_n$ comprises all the objects present across entire video frames, while the edge set $\mathcal{E}_n$ represents the bidirectional connections between objects in adjacent frames. Specifically, the graph $\mathcal{G}_n$ connects objects represented by sets $X_1, X_2, \ldots, X_k$, which are present in frames $f_1, f_2, \ldots, f_T$ of video $V_n$, through bidirectional bipartite connections. These connections are represented by the edge set $\mathcal{E}_n = B_{1 \leftrightarrow 2} \cup B_{2 \leftrightarrow 3} \cup \ldots \cup B_{(T-1) \leftrightarrow T}$. Formally, this can be expressed as:
\begin{equation}
    \mathcal{G}_n(\mathcal{V}_n, \mathcal{E}_n) = G\left(\bigcup_{i=1}^{T} X_i, \bigcup_{i=1}^{T-1} B_{i \leftrightarrow (i+1)}\right)
    \label{equ:G_n}
\end{equation}
where $B_{i \leftrightarrow (i+1)}$ represents the bidirectional bipartite edge set between objects in frames $i$ and $i+1$. The total number of nodes and edges in the graph are therefore calculated as $\sum_{i=1}^{T} |X_i|$ and $2 \sum_{i=1}^{T-1} |X_i| \times |X_{i+1}|$, respectively. We uniformly sample frames at regular intervals, specifically every 5th frame (i.e., every 0.167 seconds for a 30fps video), to avoid constructing unnecessarily large graphs. \\
\noindent \textbf{Complexity Analysis.} The complexity of constructing our $T$-layered bipartite graph $\mathcal{G}_n$ can be analyzed by evaluating the complexities associated with node and edge creation. For a video $V_n$ with an average of $\bar{x}$ objects per frame across $T$ frames, the complexity of creating the nodes is approximately $O(T\bar{x})$. The creation of edges $\mathcal{E}_n$, which involves connecting objects between adjacent frames, results in a complexity of $O(T\bar{x}^2)$. Therefore, the overall complexity of constructing the graph, combining both node and edge creation, is $O(T\bar{x}) + O(T\bar{x}^2) \approx O(T\bar{x}^2)$, assuming $\bar{x} \gg 1$. In contrast, constructing a fully connected graph with $T \times \bar{x}$ nodes would lead to a complexity of approximately $O((T\bar{x})^2)$ for large $T$ and $\bar{x}$, primarily because connections extend beyond just consecutive frames. This comparison highlights the efficiency of our bipartite graph approach in handling longer videos while keeping the average number of objects per frame constant. Notably, when $\bar{x} \approx 1$, the complexity of $\mathcal{G}_n$ simplifies to $O(T)$, indicating that the graph's complexity is more dependent on the number of frames than on the number of objects per frame.

\begin{figure}[ht]
    \vspace{-0.1cm}
    \centering
    \includegraphics[width=\linewidth]{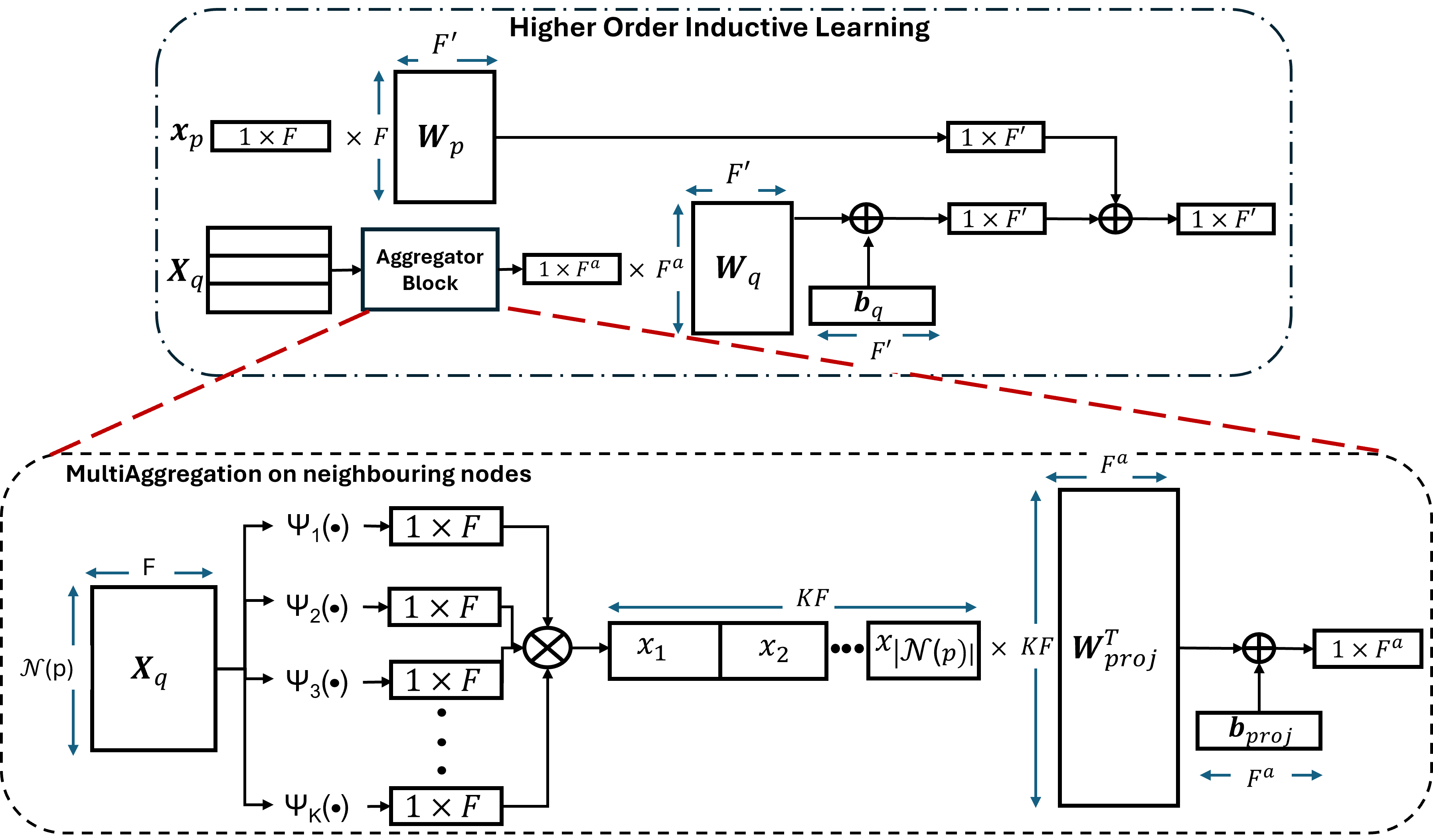}
    \caption{Illustration of the multi-aggregation process in the high-order inductive learning mechanism as discussed in Section~\ref{sub:high-order-graph-learning}. The top block shows the graph convolution where the root node is updated through transformations and aggregation. The bottom block details the statistical multi-aggregation applied to neighboring nodes, with features concatenated and projected to enhance node representations.}
    \label{fig:aggr}
    \vspace{-0.3cm}
\end{figure}

\vspace{-0.2cm}
\subsection{High-Order Inductive Learning}
\label{sub:high-order-graph-learning}
The core concept of our approach is to efficiently learn and aggregate features (illustrated in Fig.~\ref{fig:aggr}) by generating robust node embeddings from a node’s local neighborhood that generalize well to unseen interaction patterns in traffic scenes. Inspired by GraphSAGE~\cite{hamilton2017inductive}, we extend its single aggregation operation by employing a set of \(K\) trainable aggregation functions, \(\psi_k\) for each \(k \in \{1, \dots, K\}\), to aggregate information from neighbors. The aggregated features are then projected into a more informative embedding space using projection matrices \(\mathbf{W}_k\), transforming the raw features into a richer, more expressive representation.

\noindent \textbf{Generate Embedding and Update.}  
The feature information of the \(p^{th}\) root node (denoted by \(\mathbf{x}_p\)) is updated based on its neighboring nodes indexed by \(\mathcal{N}(p)\). This process begins by applying a set of \(K\) statistical aggregation functions \(\psi_k\), each independently reducing the \(|\mathcal{N}(p)| \times F\) feature matrix, constructed from the neighboring nodes, into a \(1 \times F\) vector. These aggregated features are then concatenated along the feature dimension, resulting in a final dimension of \(1 \times KF\). Without loss of generality, all the projection matrices \(\mathbf{W}_k \in \mathbb{R}^{F \times F^a}\) associated with each aggregation function can be represented as a single projection matrix \(\mathbf{W}_{proj} = [\mathbf{W}_1, \mathbf{W}_2, \ldots, \mathbf{W}_K]^T \in \mathbb{R}^{KF \times F^a}\), where \(F^a\) is the output feature dimension of an aggregation operation. Additionally, a bias term \(b_{proj} \in \mathbb{R}^{1 \times F^a}\) is added to this transformation, which can be expressed as:
\begin{equation}
    \Psi_{q \in \mathcal{N}(p)} (\mathbf{x}_q) = \mathbf{W}_{proj} \cdot \Bigl(\big\|_k \psi_k (\mathbf{x}_q)\Bigr) + \mathbf{b}_{proj},
\end{equation}
where \(q\) indexes the neighboring nodes relative to the root node. This final transformation updates the feature information of each node \(\mathbf{x}_p\) in our graph \(\mathcal{G}_n\). The update rule for the \(l^{th}\) layer of our GNN model is therefore expressed as:
\begin{equation}
    \mathbf{x}_p^{l+1} = \mathbf{W}_p^l \mathbf{x}_p^l + \mathbf{W}_q^l \cdot \Psi_{q \in \mathcal{N}(p)}^l \mathbf{x}_q^l + \mathbf{b}_q^l,
    \label{equ:sageconv}
\end{equation}
Here, \(\mathbf{W}_p \in \mathbb{R}^{F \times F'}\) and \(\mathbf{W}_q \in \mathbb{R}^{F^a \times F'}\) are separate transformations applied to the root node and the neighboring nodes, respectively, to match the desired output dimension \(F'\) of the GNN operation. \\
\noindent \textbf{Multi-Aggregation with High-Order Statistics.} 
We enhanced the aggregation process by introducing high-order statistical moments alongside the usual mean, standard deviation, and median calculations, resulting in a more robust and informative aggregation operation. Each aggregator performs an element-wise operation along each feature dimension, reducing the \(|\mathcal{N}(p)| \times F\) feature matrix \(\mathbf{X}\) into $1 \times F$ dimensional vector, where \(\mathbf{X} = [\mathbf{x}_1, \mathbf{x}_2, \dots, \mathbf{x}_{|\mathcal{N}(p)|} ]^T\) and each \(\mathbf{x}_i\) is an \(F\)-dimensional column vector. The mean aggregation function \(\psi_1(\mathbf{X}) = \mathbf{\mu} = \frac{1}{|\mathcal{N}(p)|} \sum_{i=1}^{|\mathcal{N}(p)|} \mathbf{X}_{i,:}\) computes the mean of feature $\mathbf{X}$. Similarly, the standard deviation of $\mathbf{X}$ can be calculated as \(\psi_2(\mathbf{X}) = \sigma^2 = \frac{1}{|\mathcal{N}(p)|} \sum_{i=1}^{|\mathcal{N}(p)|} (\mathbf{X}_{i,:} - \mathbf{\mu})^2\). The median aggregation function \(\psi_3\) calculates the feature-wise median. For each feature position along $|\mathcal{N}(p)|$, the median is determined by first sorting the values in ascending order and then selecting the middle value, which corresponds to the position \(i = \left\lfloor \frac{|\mathcal{N}(p)|}{2} \right\rfloor\). To extend the range of statistical properties captured by our model, we incorporate high-order statistical aggregations. These are represented as \(\psi_m(\mathbf{X}) = \frac{1}{|\mathcal{N}(p)|} \sum_{i=1}^{|\mathcal{N}(p)|} (\mathbf{X}_{i,:} - \mathbf{\mu})^m\), where \(m\) denotes the order of the statistic. Specifically, we use high-order aggregations with \(m = 3\) and \(m = 4\), in addition to the mean, standard deviation, and median. This multi-aggregation strategy ensures that our model captures a diverse range of statistical properties from the node’s local neighborhood, leading to a more comprehensive and expressive node embedding.

\noindent \textbf{Feature-Gated Attention Pooling.} 
After generating the output features from the final layer of our GNN model, we apply feature-gated attention pooling to aggregate node features across the graph \( \mathcal{V}_n \) and produce the final prediction for the video sequence. This mechanism selectively emphasizes the most relevant nodes, enhancing the model's ability to capture critical aspects of the video. Let the output of the final GNN layer be \(\hat{\mathbf{X}} \in \mathbb{R}^{|\mathcal{V}_n| \times \hat{F}}\), where \( |\mathcal{V}_n| \) is the number of nodes in the graph corresponding to the video \( V_n \). The feature-gated attention pooling is expressed as:
\begin{equation}
    \mathbf{X}_{pooled} = \sum_{i=1}^{|\mathcal{V}_n|} \mathcal{S} \Bigl(\hat{\mathbf{X}}_i \mathbf{W}_1 + \mathbf{b}_1 \Bigr) \odot \hat{\mathbf{X}}_i
\end{equation}
where term \(\mathcal{S} (\cdot) = e^{(\cdot)} / \sum^{|\mathcal{V}_n|} e^{(\cdot)}\) is a softmax operation applied over \(\hat{\mathbf{X}}_i\) using a learned transformation with weight matrices \( \mathbf{W}_1 \in \mathbb{R}^{\hat{F} \times \hat{F}}\) and bias \( \mathbf{b}_1 \in \mathbb{R}^{1 \times \hat{F}} \). This serves as a feature-gated attention mechanism, modulating the importance of each node’s features through an element-wise Hadamard product (denoted by \( \odot \)) with the input matrix \(\hat{\mathbf{X}}\). By gating features, the pooling operation prioritizes the most relevant nodes, ensuring the model focuses on those contributing most to the action representation. The pooled output \(\mathbf{X}_{pooled}\) is then passed through a fully connected layer followed by a softmax activation function to yield the final prediction \(\hat{p}_n\) for the video sequence \(V_n\).

\vspace{-0.3cm}
\section{Experiments}
\label{sec:exp}
\vspace{-0.1cm}
In this section, we explore the datasets utilized, the evaluation metrics applied, and the implementation specifics. Following this, we present a performance comparison with existing SotA methods and conclude with an in-depth ablation analysis.
\vspace{-0.2cm}
\subsection{Dataset Description and Evaluation}
Our method was evaluated using two primary datasets: the ROad event Awareness Dataset (ROAD) Dataset~\cite{singh2022road} and the ROAD Waymo Dataset~\cite{ettinger2021large}. The ROAD is specifically tailored for autonomous driving research, comprising 122,000 frames extracted from 22 videos, each approximately 8 minutes long. The dataset was collected using a camera mounted on an autonomous vehicle (AV). It features two distinct sets of action labels. The first set, known as agent-level actions, describes the behaviors of various road agents (e.g., pedestrians, vehicles, cyclists) in a scene, such as ``waiting2Cross'' or ``overtaking'', particularly in static contexts where the AV is not in motion. The second set, referred to as AV-actions, characterizes the autonomous vehicle's own movements, independent of the actions of other agents in the scene, with labels such as ``AV-on-the-move'' and ``AV-stopped''. The ROAD Waymo Dataset builds upon the ROAD dataset by incorporating data from the Waymo Open Dataset~\cite{ettinger2021large}. It consists of 198,000 annotated frames from 1,000 videos, each averaging 20 seconds in duration. This dataset offers action annotations for various road agents, along with agent type and semantic location labels. It contains 54,000 tracks, with over 3.9 million bounding box-level agent labels and more than 4.2 million action and location labels. The ROAD Waymo Dataset is designed for the robust evaluation of autonomous driving models, particularly in complex and diverse driving scenarios captured by Waymo's autonomous vehicles. For evaluation, both datasets provide predefined splits for training, validation, and testing. Model performance was assessed using classification accuracy (Acc. in \%) and mean Average Precision (mAP in \%).

\vspace{-0.1cm}
\subsection{Implementation Details}
\vspace{-0.1cm}
Our implementation is built using PyTorch 2.12 and PyTorch Geometric 2.1. The proposed model architecture consists of two GNN layers that leverage multi-aggregation of high-order statistics from neighboring nodes to effectively process graph data. For each input video, we first extract spatio-temporal tube regions based on the bounding boxes of objects in the video frames. These regions are cropped across \(\tau = 16\) frames, centered at a middle frame as described in Section~\ref{sub:spatio-temporal-tubes}. To ensure that the bounding boxes encompass the entire object appearance over the 16 frames, the bounding boxes are scaled by a factor of 1.5 before cropping. Each object instance is then resized to a fixed dimension of \(256 \times 256\), constructing a spatio-temporal tube represented as a 4D tensor of size \(16 \times 256 \times 256 \times 3\). These tensors are fed into a Kinetics pre-trained 3D CNN feature extractor, specifically the S3D model~\cite{xie2018rethinking}, which outputs a 1024-dimensional vector for each spatio-temporal tube. These vectors serve as nodes in our constructed graph, with each node representing a distinct object. The node features are then processed through our proposed high-order statistics-based multi-aggregation GNN. To achieve a more meaningful and informative representation, we employ a bottleneck mechanism between the two GNN layers. In the first GNN layer, the input node feature dimension is reduced by half to 512. The output from this layer is then passed through a second GNN layer, where the feature dimension is restored to 1024. Next, we apply feature-gated attention pooling, as described in Section~\ref{sub:high-order-graph-learning}, directly to the 1024-dimensional output features. Since we use element-wise multiplication in the gating mechanism, the output dimension of the transformation weights is kept consistent at 1024, matching the output feature dimension. Our GNN model is trained using the Adam optimizer with a learning rate of $1 \times 10^4$, a weight decay of 0.5, a batch size of 32 sequences, and for 200 epochs. During ablation studies with different backbone networks, which may have varying input node feature dimensions, we maintain the bottleneck feature dimension between the two GNN layers at a ratio of 0.5.

\begin{table}[ht!]
\vspace{-0.3cm}
\centering
\caption{Comparison of the results using features from different feature extractors on the agent tube features. The best results are shown in \textbf{bold}.}
\begin{tabular}{|c|cc|cc|cc|cc|}
\hline
 \multirow{2}{*}{\textbf{Agent Tube Features}} & \multicolumn{2}{c|}{\textbf{S3D}~\cite{xie2018rethinking}} & \multicolumn{2}{c|}{\textbf{R(2+1)D}~\cite{tran2018closer}} & \multicolumn{2}{c|}{\textbf{SlowFast}~\cite{feichtenhofer2019slowfast}} & \multicolumn{2}{c|}{\textbf{3D RES.}~\cite{hara2018towards}} \\ 
\cline{2-9}
& \textbf{Acc.} & \textbf{mAP} & \textbf{Acc.} & \textbf{mAP} & \textbf{Acc.} & \textbf{mAP} & \textbf{Acc.} & \textbf{mAP} \\ 
\hline
 ROAD Dataset & \textbf{55.87} & \textbf{34.43} & 60.42 & 34.80 & 56.44 & 30.26 & 47.69 & 19.26 \\ 
 ROAD Waymo Dataset & \textbf{57.39}  & \textbf{21.30}  & 59.90 & 23.21  & 58.05  & 19.86  & 52.34 & 14.74 \\ 
\hline
\end{tabular}
\label{tab:tube_features}
\vspace{-0.7cm}
\end{table}
\vspace{-0.3cm}
\subsection{Performance Comparison}
\vspace{-0.1cm}
In this work, we establish a baseline by approaching the problem as a classification task on the ROAD and ROAD Waymo datasets, utilizing various video processing backbones. Table \ref{tab:tube_features} compares the performance of four feature extraction methods like S3D~\cite{xie2018rethinking}, R(2+1)D~\cite{tran2018closer}, SlowFast~\cite{feichtenhofer2019slowfast}, and 3D ResNet~\cite{hara2018towards} on the ROAD and ROAD Waymo datasets, highlighting the effectiveness of our high-order GNN approach across different off-the-shelf backbones. R(2+1)D demonstrates the highest accuracy on both datasets, with 60.42\% on ROAD and 59.90\% on ROAD Waymo, showcasing its strong ability to correctly identify actions in driving scenes. However, its mAP scores, which measure the precision of ranking detections, decline significantly on the ROAD Waymo dataset (34.80\% on ROAD and 23.21\% on ROAD Waymo). This indicates that while R(2+1)D is effective at recognizing actions, it struggles with accurately ranking them in more complex scenarios. S3D provides a more balanced performance, achieving accuracies of 55.87\% on ROAD and 57.39\% on ROAD Waymo, with mAPs of 34.43\% and 21.30\%, respectively. Although S3D's accuracy is slightly lower than R(2+1)D's, its more consistent mAP scores suggest it offers better precision across various actions, making it a reliable choice when both accuracy and precise ranking are important. SlowFast performs moderately well, with accuracies of 56.44\% and 58.05\%, but its mAP scores (30.26\% and 19.86\%) suggest it is less effective at accurately ranking detected actions. This may be due to its dual-pathway design, which might not fully align with the specific dynamics of these datasets. 3D ResNet shows the lowest performance, with accuracies of 47.69\% on ROAD and 52.34\% on ROAD Waymo, and mAPs of 19.26\% and 14.74\%. These results highlight its challenges in both identifying and accurately ranking actions in complex driving scenes. In summary, R(2+1)D is most effective for high accuracy in action identification, while S3D offers a balanced approach with reliable precision in action ranking. SlowFast and 3D ResNet may require further optimization to better handle the complexities of these driving scenarios. \\
\begin{table}[ht!]
\vspace{-0.4cm}
\centering
\caption{Performance comparison of S3D and R(2+1)D features computed for AV-actions for ROAD and ROAD Waymo datasets.}
\renewcommand{\arraystretch}{1.2}
\setlength{\tabcolsep}{6pt}
\begin{adjustbox}{max width=\textwidth}
\begin{tabular}{|c|cc|cc|}
\hline
\multirow{2}{*}{\textbf{AV}  \textbf{Features} } & \multicolumn{2}{c|}{\textbf{S3D}~\cite{xie2018rethinking}} & \multicolumn{2}{c|}{\textbf{R(2+1)D}~\cite{tran2018closer}} \\ 
\cline{2-5}
& \textbf{Acc.} & \textbf{mAP} & \textbf{Acc.} & \textbf{mAP} \\ 
\hline
\textbf{ROAD} & 87.91 & 41.71 & 87.93 & 40.14 \\ 
\hline
\textbf{ROAD Waymo} & 93.71  & 36.16  & 93.14 & 30.42 \\ 
\hline
\end{tabular}
\end{adjustbox}
\label{tab:av_features}
\vspace{-0.4cm}
\end{table}
Table \ref{tab:av_features} provides a comparison of action classification performance using Autonomous Vehicle (AV) features, focusing on the interactions between multiple objects and the overall scene dynamics, across the ROAD and ROAD Waymo datasets. On the ROAD dataset, both S3D and R(2+1)D models perform exceptionally well, achieving nearly identical accuracies of 87.91\% and 87.93\%, respectively. However, S3D slightly outperforms R(2+1)D in terms of mAP, with scores of 41.71\% versus 40.14\%. This suggests that while both models are adept at recognizing driving actions, S3D is marginally better at ranking the detected actions, which could be crucial in scenarios requiring precise action ordering. On the ROAD Waymo dataset, which presents more complex and varied scenes, S3D continues to lead with an accuracy of 93.71\% and an mAP of 36.16\%, compared to R(2+1)D's 93.14\% accuracy and 30.42\% mAP. The decrease in mAP for both models on this dataset underscores the challenge of accurately ranking actions in more diverse environments. Nevertheless, S3D's superior mAP suggests it remains more reliable for tasks where the accurate ranking of detected actions is critical. Overall, these results demonstrate that S3D consistently provides better action ranking, particularly in complex environments, making it a strong candidate for scenarios requiring precision in both identification and ranking. R(2+1)D remains highly accurate in action identification, which is beneficial in contexts where correct detection is paramount.

\begin{table}[h!]
\vspace{-0.1cm}
\centering
\caption{Accuracies and mAP for agent-level and AV-level actions on ROAD and ROAD Waymo datasets, using different aggregation combinations with S3D~\cite{xie2018rethinking} and R(2+1)D~\cite{tran2018closer} backbones. The best results are shown in \textbf{bold}.}
\setlength{\tabcolsep}{2.5pt} 
\renewcommand{\arraystretch}{0.95} 
\begin{adjustbox}{max width=\textwidth}
\begin{tabular}{|c|ccccc|cc|cc|cc|cc|}
\hline
\multirow{3}{*}{\textbf{Dataset}} & \multicolumn{5}{c|}{\textbf{Aggregation combination(s)}} & \multicolumn{4}{c|}{\textbf{S3D}~\cite{xie2018rethinking}} & \multicolumn{4}{c|}{\textbf{R(2+1)D}~\cite{tran2018closer}} \\ 
\cline{2-14}
& \multirow{2}{*}{\textbf{Mean}} & \multirow{2}{*}{\textbf{Med.}} & \multirow{2}{*}{\textbf{Std}} & \multirow{2}{*}{\textbf{m=3}} & \multirow{2}{*}{\textbf{m=4}} & \multicolumn{2}{c|}{\textbf{Agent}} & \multicolumn{2}{c|}{\textbf{AV}} & \multicolumn{2}{c|}{\textbf{Agent}} & \multicolumn{2}{c|}{\textbf{AV}} \\ 
 &  &  &  &  &  & \textbf{Acc.} & \textbf{mAP} & \textbf{Acc.} & \textbf{mAP} & \textbf{Acc.} & \textbf{mAP} & \textbf{Acc.} & \textbf{mAP} \\ 
\hline
\multirow{5}{*}{\textbf{ROAD}} & \checkmark &  &  &  &  & 54.60  & 27.54  & 85.73  & 38.86 & 59.04  & 33.22  & 85.67  & 38.40  \\ 
 & \checkmark & \checkmark &   &  &  & 54.90  & 28.56 & 86.08 & 39.90  & 59.35  & 33.35  & 85.97  & 38.76  \\ 
 & \checkmark & \checkmark & \checkmark &  &  &55.82  &28.70  & 86.20  & 40.95  & 59.50  & 33.94 & 86.12  & 38.84 \\ 
 & \checkmark & \checkmark & \checkmark & \checkmark &  & 55.45 & 30.31  & 86.79  & 41.35 & 60.27  & 34.20  & 86.44   & 39.63  \\ 
 & \checkmark & \checkmark & \checkmark & \checkmark & \checkmark  & \textbf{55.87}  & \textbf{34.43} & \textbf{87.91} & \textbf{41.71} & \textbf{60.42} & \textbf{34.80} & \textbf{87.93}   & \textbf{40.14} \\
\hline
 \multirow{5}{*}{\shortstack{\textbf{Road} \\ \textbf{Waymo}}} & \checkmark &   &  &  &  & 52.97  & 18.78  & 92.80 & 30.39   & 56.91 & 21.78  & 90.53 & 28.26 \\ 
& \checkmark & \checkmark &  &  &  & 53.23 & 19.56 & 93.04 & 31.26  &56.34  &21.41  & 92.51  & 29.94  \\ 
& \checkmark & \checkmark & \checkmark &  &  & 54.17 & 19.62   & 93.05  & 35.58  & 55.31  & 21.96 &92.76  & 30.34 \\ 
 & \checkmark & \checkmark & \checkmark & \checkmark &  & 54.53  & 20.56  & 93.48 & 35.65  & 56.28 & 22.71  & 92.80   & 30.39  \\ 
 & \checkmark & \checkmark & \checkmark & \checkmark & \checkmark  & \textbf{57.39} & \textbf{21.30}  & \textbf{93.71}  & \textbf{36.16}  & \textbf{59.90}  & \textbf{23.21}  & \textbf{93.14}  & \textbf{30.42}  \\
\hline
\end{tabular}
\end{adjustbox}
\label{tab:aggregation_combinations}
\vspace{-0.4cm}
\end{table}
\vspace{-0.1cm}
\subsection{Ablation Studies}
\vspace{-0.1cm}
\textbf{Impact of varying aggregations.}
Table \ref{tab:aggregation_combinations} analyzes the impact of different aggregation strategies within our GNN on the performance of action recognition in the ROAD and ROAD Waymo datasets, covering both agent-level and AV-level actions. The results show that applying multiple aggregation techniques significantly improves performance for both S3D and R(2+1)D feature extractors. For agent-level actions, both S3D and R(2+1)D feature extractors show notable improvements when multiple aggregation techniques are applied. On the ROAD dataset, R(2+1)D achieves its highest accuracy of 60.42\% and mAP of 34.80\% with a combination of mean, median, standard deviation, and high-order moments (m=3 and m=4). This combination allows the model to effectively capture central tendencies, variability, and subtle patterns within the spatio-temporal features. S3D also benefits from this approach, achieving an accuracy of 55.87\% and mAP of 34.43\%, highlighting the importance of diverse statistical measures in enhancing node representations. A similar trend is observed for AV-level actions. The most effective aggregation strategy for S3D, incorporating mean, median, standard deviation, and high-order moments, results in an accuracy of 87.91\% and mAP of 41.71\% on the ROAD dataset. This approach helps the GNN capture intricate patterns and subtle variations in the data, crucial for distinguishing different actions in complex traffic scenarios. On the ROAD Waymo dataset, the benefits of these aggregation strategies are even more evident. S3D achieves its highest performance with an accuracy of 57.39\% and mAP of 21.30\% for agent-level actions, and 93.71\% accuracy with 36.16\% mAP for AV-level actions. R(2+1)D similarly benefits, reaching 59.90\% accuracy and 23.21\% mAP for agent-level actions, and 93.14\% accuracy with 30.42\% mAP for AV-level actions. The use of these diverse aggregation strategies within the GNN framework enhances the model’s ability to represent spatio-temporal features and manage the complexities of dynamic driving scenarios. By capturing a broad range of statistical properties such as central tendencies, variability, and high-order patterns—the model is better equipped to differentiate between various actions and interactions, leading to improved overall performance in both agent-level and AV-level contexts.

\begin{table}[h!]
\vspace{-0.1cm}
\centering
\caption{Impact of different pooling mechanisms on agent-level and AV-level actions on ROAD datasets across varying architectures. The results are shown in \textbf{bold.}}
\renewcommand{\arraystretch}{1.2}
\setlength{\tabcolsep}{4pt}
\begin{adjustbox}{max width=\textwidth}
\begin{tabular}{|c|cc|cc|cc|cc|}
\hline
\multirow{3}{*}{\shortstack{\textbf{Pooling} \\ \textbf{mechanism}}} & \multicolumn{4}{c|}{\textbf{S3D}~\cite{xie2018rethinking}} & \multicolumn{4}{c|}{\textbf{R(2+1)D}~\cite{tran2018closer}} \\ 
\cline{2-9}
& \multicolumn{2}{c|}{\textbf{Agent}} & \multicolumn{2}{c|}{\textbf{AV}} & \multicolumn{2}{c|}{\textbf{Agent}} & \multicolumn{2}{c|}{\textbf{AV}} \\ 
\cline{2-9}
 & \textbf{Acc.} & \textbf{mAP} & \textbf{Acc.} & \textbf{mAP} & \textbf{Acc.} & \textbf{mAP} & \textbf{Acc.} & \textbf{mAP} \\ 
\hline
\textbf{Global Mean} & 51.53  & 27.38 & 84.78  & 40.20 & 53.22  & 24.78  & 85.37  & 38.83 \\  
\textbf{Global Sum} & 54.14  & 29.44  & 86.08 & 40.65   & 54.29 & 27.15  & 85.82  & 38.30  \\  
\textbf{Global Max} & 49.23 & 26.38  & 84.08 & 39.30 & 56.74 & 30.36  & 85.67  & 37.85  \\  
\textbf{Feature-Gated Attn. (Ours)} & \textbf{55.87}  & \textbf{34.43} & \textbf{87.91} & \textbf{41.71}  & \textbf{60.42}  & \textbf{34.80}  & \textbf{87.93}  & \textbf{40.14}  \\ 
\hline
\end{tabular}
\end{adjustbox}
\label{tab:key_modules}
\vspace{-0.7cm}
\end{table}

\noindent \textbf{Impact of different pooling mechanisms.}
The results in Table \ref{tab:key_modules} demonstrate the effectiveness of various pooling and aggregation methods in traffic scene analysis, particularly on the ROAD dataset. In these scenarios, the choice of pooling mechanism is crucial for capturing relevant spatio-temporal patterns, which directly impacts the model’s ability to interpret dynamic interactions. Global Mean pooling averages features across nodes, tends to oversimplify the diverse interactions in traffic scenes. This simplification often results in the loss of important information, as reflected in the lower performance metrics (e.g., 51.53\% accuracy and 27.38\% mAP for Agent-level action using S3D). Global Add pooling operation is more informative by summing features, can still miss subtle yet important variations in the data, leading to only modest improvements (e.g., 54.29\% accuracy and 27.15\% mAP for Agent-level action using R(2+1)D). Global Max pooling operation is more effective at capturing the most prominent features, fails to consider the nuanced interactions essential in complex environments like traffic scenes. By focusing solely on maximum values, this method overlooks the broader context of traffic dynamics, resulting in subpar performance (e.g., 49.23\% and 56.74\% for Agent-level action using S3D and R(2+1)D respectively). In contrast, our Feature-Gated Attention (Attn.) mechanism shows superior performance by dynamically prioritizing the most relevant features within the graph. This method is particularly beneficial in traffic scene analysis, where multiple agents interact in complex ways that require careful consideration of context and relevance. By selectively attending to critical features, Feature-Gated Attention significantly enhances the model’s ability to distinguish subtle variations in traffic behaviors and interactions. This leads to much higher accuracy and mAP scores across both S3D and R(2+1)D architectures (\eg, 55.87\% accuracy and 34.43\% mAP  for Agent-level action using S3D, 87.91\% accuracy and 41.71\% mAP  for AV-level action using S3D). This ablation highlights the effectiveness of attention mechanisms in GNNs for traffic scene analysis, where accurately capturing the intricate dynamics of road users is crucial for robust action recognition.

\vspace{-0.2cm}
\begin{table}[th!]
\vspace{-0.2cm}
\centering
\caption{Impact of feature dimension bottlenecking (Compression) between two GNN layers on performance across different datasets and architectures.}
\renewcommand{\arraystretch}{1.2}
\setlength{\tabcolsep}{3pt}
\begin{adjustbox}{max width=\textwidth}
\begin{tabular}{|c|cc|cc|cc|cc|cc|cc|cc|cc|cc|}
\hline
  \multirow{4}{*}{\shortstack{\textbf{GNN Intermediate} \\ \textbf{Feat. Dim.} }}  & \multicolumn{8}{c|}{\textbf{S3D}~\cite{xie2018rethinking}} & \multicolumn{8}{c|}{\textbf{R(2+1)D}~\cite{tran2018closer}} \\ 
\cline{2-17}
& \multicolumn{4}{c|}{\textbf{Road}} & \multicolumn{4}{c|}{\textbf{Road Waymo}} & \multicolumn{4}{c|}{\textbf{Road}} & \multicolumn{4}{c|}{\textbf{Road Waymo}} \\ 
\cline{2-17}
& \multicolumn{2}{c|}{\textbf{Agent}} & \multicolumn{2}{c|}{\textbf{AV}} & \multicolumn{2}{c|}{\textbf{Agent}} & \multicolumn{2}{c|}{\textbf{AV}} & \multicolumn{2}{c|}{\textbf{Tube}} & \multicolumn{2}{c|}{\textbf{AV}} & \multicolumn{2}{c|}{\textbf{Tube}} & \multicolumn{2}{c|}{\textbf{AV}} \\ 
\cline{2-17}
 & \textbf{Acc.} & \textbf{mAP} & \textbf{Acc.} & \textbf{mAP} & \textbf{Acc.} & \textbf{mAP} & \textbf{Acc.} & \textbf{mAP} & \textbf{Acc.} & \textbf{mAP} & \textbf{Acc.} & \textbf{mAP} & \textbf{Acc.} & \textbf{mAP} & \textbf{Acc.} & \textbf{mAP} \\ 
\hline
\textbf{W/o Compression} & 48.62  & 23.83  & 82.31  & 37.33  & 54.53  & 20.57  & 93.69 & 36.10  & 54.29  & 27.31  & 83.57 & 36.14 & 53.63 & 20.28  & 92.74   & 29.76   \\ 
\hline
\textbf{W Compression (Ours)} & 55.87 & 34.43  & 87.91   & 41.71  & 57.39   & 21.30 &93.71  & 36.16 & 60.42 & 34.80  &87.93  & 40.14  & 59.91  & 23.21  & 93.14  & 30.42  \\ 
\hline
\end{tabular}
\end{adjustbox}
\label{tab:bottlenecking}
\vspace{-0.3cm}
\end{table}
\vspace{-0.2cm}
\noindent \textbf{Impact of feature compression.} 
The results in Table \ref{tab:bottlenecking} highlight the significant performance improvements achieved by applying feature compression (bottlenecking) between GNN layers across different datasets and architectures. For example, in the ROAD dataset using the S3D architecture, feature compression boosted the accuracy for Agent actions from 48.62\% to 55.87\% and mAP from 23.83\% to 34.43\%. Similarly, in the ROAD Waymo dataset, the accuracy and mAP for Agent actions increased from 54.53\% to 57.39\% and from 20.57\% to 21.30\%, respectively. These improvements are consistent across both S3D and R(2+1)D architectures and various action types, including agent and AV level actions, demonstrating that feature compression consistently enhances model performance. Technically, this enhancement is due to the bottleneck enforcing more efficient and discriminative feature representations, similar to the compression phase in autoencoders. By reducing the feature dimensions passed between layers, the GNN focuses on capturing the most critical aspects of the input data, which is vital for accurately recognizing complex spatio-temporal dynamics in traffic scenes. Moreover, feature compression helps mitigate overfitting, particularly with large and noisy datasets, by preventing the model from learning irrelevant correlations. Overall, applying feature compression between GNN layers results in a more efficient and robust model, improving generalization and performance in challenging environments like traffic scenes.

\vspace{-0.2cm}
\section{Conclusion}
\vspace{-0.1cm}
In conclusion, this paper presents a comprehensive framework that leverages High-Order Evolving Graphs and advanced GNNs to enhance the modeling and representation of traffic dynamics in autonomous driving applications. Our approach integrates temporal bidirectional bipartite graphs with 3D CNN features, multi-aggregation strategies, and attentional mechanisms as part of inductive learning. This combination effectively captures complex spatio-temporal interactions within dynamic traffic scenes, significantly improving both the accuracy and detail in traffic scene analysis. The effectiveness of our method is demonstrated through extensive experiments on the ROAD and ROAD Waymo datasets, where it establishes a comprehensive baseline for further developments in traffic behavior analysis. The results highlight the importance of integrating detailed spatio-temporal features with advanced graph-based techniques to achieve accurate and scalable traffic behavior analysis.

\noindent \textbf{Acknowledgement:}
This research was partially funded by the UKRI EPSRC project ATRACT (EP/X028631/1): A Trustworthy Robotic Autonomous System for Casualty Triage.
\vspace{-0.4cm}
%
%
\bibliographystyle{splncs04}
\bibliography{main}
\end{document}